\documentclass[letterpaper, 10 pt, conference]{ieeeconf}
\IEEEoverridecommandlockouts                   
\usepackage{cite}
\usepackage{amsmath,amssymb,amsfonts}
\usepackage{graphicx}
\usepackage{textcomp}
\usepackage{xcolor}
\usepackage{mathabx}
\usepackage{makecell}
\usepackage{subfig}

\usepackage{algorithmicx}
\usepackage[noend]{algpseudocode}
\usepackage{algorithm} 
\usepackage{booktabs}
\usepackage{hyperref}

\usepackage{balance}
\usepackage{lastpage}
\DeclareMathOperator*{\argmin}{argmin}
\DeclareMathOperator*{\argmax}{argmax}

\newcommand{\norm}[1]{\left\lVert#1\right\rVert}

\usepackage{amsthm}
\newtheorem{remark}{Remark}
\newtheorem{theorem}{Theorem}
\usepackage{url}
\usepackage{mathtools}

\title{\LARGE \bf
Vector Cost Behavioral Planning for Autonomous Robotic Systems \\with Contemporary Validation Strategies
}

\makeatletter
\def\IEEEauthorrefmark#1{\textsuperscript{#1}}
\makeatother

\author{Benjamin R. Toaz\IEEEauthorrefmark{1},
        Quentin Goss\IEEEauthorrefmark{2},
        John Thompson\IEEEauthorrefmark{2}, \\
        Seta Bo\u{g}osyan\IEEEauthorrefmark{3},
        Shaunak D. Bopardikar\IEEEauthorrefmark{1},
        Mustafa \.{I}lhan Akba\c{s}\IEEEauthorrefmark{2}, 
        Metin G\"{o}ka\c{s}an\IEEEauthorrefmark{4}%
\thanks{This work was funded in part by the National Science Foundation International Research Experience for Students award number 2246347, the National Science Foundation CAREER Award ECCS-2236537, and supported by the Spartan Autonomous Racing Club at Michigan State University.}
\thanks{Special thanks to Mehmet Emre Demir for his input on the practical implementation aspects.}
\thanks{\IEEEauthorrefmark{1}\textit{Dept. of Electrical Engineering}, \textit{Michigan State University}, Lansing, Michigan, USA, {\tt\small toazbenj@msu.edu, shaunak@egr.msu.edu}}%
\thanks{\IEEEauthorrefmark{2} \textit{Dept. of Electrical Engineering \& Computer Science}, \textit{Embry-Riddle Aeronautical University}, Daytona Beach, Florida, USA, \{gossq, thomj130\}@my.erau.edu, akbasm@erau.edu}%
\thanks{\IEEEauthorrefmark{3} \textit{Dept. of Computer Science}, City College of New York, New York, New York, USA, and Halic University, Istanbul, Turkey
{\tt\small sbogosyan@alaska.edu}}%
\thanks{\IEEEauthorrefmark{4} \textit{Dept. of Electrical and Electronics Engineering}, \textit{Istanbul Technical University}, Istanbul, Turkey, {\tt\small gokasan@itu.edu.tr}}%
}

\begin{document}

\maketitle
\thispagestyle{empty}
\pagestyle{empty}

\begin{abstract}

The vector cost bimatrix game is a method for multi-objective decision making that enables autonomous robotic systems to optimize for multiple goals at once while avoiding worst-case scenarios in neglected objectives. We expand this approach to arbitrary numbers of objectives and compare its performance to scalar weighted sum methods during competitive motion planning. Explainable Artificial Intelligence (XAI) software is used to aid in the analysis of high dimensional decision-making data. 
%
%
 %
State-space Exploration of Multidimensional Boundaries using Adherence Strategies (SEMBAS) is applied to explore performance modes in the parameter space as a sensitivity study for the baseline and proposed frameworks. 
%
%
While some works have explored aspects of game theoretic planning and intelligent systems validation separately, we combine each of these into a novel and comprehensive simulation pipeline. This integration demonstrates a dramatic improvement of the vector cost method over scalarization and offers an interpretable and generalizable framework for robotic behavioral planning. Code available at \url{https://github.com/toazbenj/race_simulation}.
The video companion to this work is available at \url{https://tinyurl.com/vectorcostvideo}.\end{abstract} 

\section{INTRODUCTION}


Decision-making by intelligent agents can be modeled as a game in which multiple players compete for and share resources. This could include control of a plant in order to manipulate an object, nodes in a network that alter the flow of information, or space on a road and key positions in traffic. Each player receives different outcomes as a result of both their own actions and the actions of other players, a type of exchange called a bimatrix game. When applying these abstract approximations of decision-making to concrete scenarios with physical properties, there is often a disconnect between the assumptions made in the model and how the system performs in reality \cite{Sell2022Mar}. 

In addition to the labor-intensive task of finding useful failure cases at all, it is difficult to identify which factors are responsible for limiting a method's generalizability. There are often many interconnected causes, which makes the quantification of their effects a non-intuitive task~\cite{mohseni2020}. This challenge is further complicated by a lack of systematic procedures and tools for evaluating the disparity between game-theoretic models and the mechanics of practical systems.

These challenges are persistent within the domain of autonomous vehicles (AVs), unmanned systems in general, and in validation and verification (V\&V) testing of these complex technologies \cite{Malayjerdi2023Aug}. Structured V\&V frameworks for autonomous systems enable standardization, promoting transparency and shareability of AV research, including applications in behavioral planning games \cite{Luckcuck2019}. These frameworks, when complemented with XAI and boundary exploration techniques, also provide concise graphical interpretations and explanations throughout all stages of research and development, which are essential for reporting, debugging, and fostering user and public acceptance \cite{Omeiza2022Explanations}. Our objective for this work is to facilitate the integration of these tools for V\&V testing of game-theoretic decision making and multi-agent strategy for autonomous robots.     

\subsection{Related Literature}


Zero-sum games are simple interactions where two players operate in direct opposition to pursue a single goal \cite{NoncooperativeGame}. To capture the structure of behavioral planning for autonomous vehicles, the zero-sum game has been augmented with detailed safety constraints external to the game itself or by composing detailed cost functions from multiple metrics \cite{EncodingDefensive}, \cite{GameTheoreticPlanning}, \cite{StochasticGames}. Extensions using bimatrix games do not follow this simple structure and can also capture the cooperative parts of motion planning, though still with a single composite objective function \cite{ANoncooperativeGame}, \cite{PotentialGameBased}. These combined functions often optimize certain objectives at the expense of others and are less robust to worst-case scenarios. 

Considering more than one objective at once is a more complex and expensive approach, but can often find the most desirable tradeoffs between conflicting goals, or outcomes that lie on the Pareto frontier \cite{ScalarizingMultiObjective}. Vector cost games can be thought of as playing multiple scalar cost games at once, with the goal of arriving at a Pareto optimal outcome with respect to each sub-game \cite{GamesWithVector}, \cite{EquilibriumPoints}. Some of the multi-objective aspects of this framework have been applied to racing games for autonomous vehicles \cite{BridgingTheGap}. Deciding which Pareto front solutions to target as an autonomous agent is a subtle problem that can be solved using optimization techniques \cite{VectorCostBimatrix}. These vector cost methods can outperform scalarization approaches using the weighted sum of costs by preserving the structure of each individual cost function. This structure is also useful for ensuring outcome guarantees with respect to every objective.


Scenario-based V\&V testing is a strategy for evaluating complex systems, particularly AVs. It centers on scenarios in which an autonomous system is placed in a realistic environment, assigned a goal or mission, and observed as the scenario unfolds \cite{dreossi2019verifai}.
Prominent scenario-based AV V\&V frameworks include Foretify \cite{foretify_new}, Scenic \cite{Fremont2023Oct_Scenic2}, and Polyverif \cite{Razdan2023_PolyVerif}. These frameworks provide toolkits or software pipelines that support the definition, formatting, selection, execution, and analysis of AV V\&V scenarios.
Their robustness and success stem largely from formal scenario specification, implemented in accordance with ASAM OpenSCENARIO v1.2 \cite{ASAM-OpenSCENARIO-1.2} or v2.0 \cite{ASAM-OpenSCENARIO-2.0} best practices. This ensures that AV V\&V scenarios are both reproducible and shareable.

Rigid frameworks often fail to address specific use-case requirements, prompting the integration of strategies from multiple frameworks.
A key limitation arises from time and resource constraints, which restrict the total number of scenario tests that can be executed. This limitation can be mitigated through intelligent scenario selection strategies that prioritize high-value scenarios to optimize testing resources, in contrast to the Monte Carlo–based approaches employed by Foretify \cite{Goss2021modular}, Scenic \cite{Goss2023SUMO-SCENIC}, and Polyverif  \cite{Akbas2023_polyverif}.
Additional tools for scenario-based AV V\&V include the Boundary Adherence Strategy \cite{Goss2024Integrated,Goss2025Integrated} and SEMBAS \cite{SEMBAS}, both of which exploit topological information from the testing process to identify edge scenarios.

Another driver for integrating additional tools is scenario interpretation and explanation, a challenging task given the scale and sparsity of scenario test data \cite{Kuznietsov2024Explainable}. Intelligent strategies address this by providing graphical and non-graphical interpretations via topological analysis of performance clusters \cite{Goss2022eagle,Mullins2018Mar,Neelofar2024Apr} and by applying explainable artificial intelligence (XAI) techniques such as SHAP \cite{Goss2024Integrated,Goss2025Integrated,Liu2023understanding,tahir2024novel}, LIME \cite{Goss2025Integrated,rjoub2023explainable,tahir2024novel}, and LightGBM explanations \cite{Goss2025Integrated,ke2017lightgbm}.
To the best of our knowledge, this work represents the first application of SHAP-based XAI to the analysis of behavioral planning games in robotics.

Prior work on scenario-based validation of autonomous systems has explored a wide range of methods for navigating high-dimensional parameter spaces. Global sampling approaches have been used for unmanned vehicle missions to identify areas of interest in three dimensions~\cite{Mullins2018Mar}, while probabilistic scenario definition has been applied to generate diverse autonomous driving cases in higher dimensions~\cite{Goss2022eagle}. Other methods such as Optimization Searching (OS)~\cite{Zhu2021hazardous} attempt to explicitly distinguish hazardous from safe regions to bias testing toward safety-critical scenarios. Boundary-focused approaches also exist: surrogate models and gradient descent have been used to estimate ``safety performance boundaries''~\cite{wang2022safety}, while Gaussian Processes~\cite{akellaa2022sample} offer probabilistic predictions of safe/unsafe divisions. Path planning methods such as RRT and its extensions~\cite{Noreen2016Nov, Xanthidis2016RRTF} have similarly been adapted to explore parameter configurations in high dimensions. However, these methods typically rely on surrogate models or post-processing steps to infer boundaries, making them dependent on additional training or clustering. In contrast, SEMBAS~\cite{SEMBAS} directly samples along the performance boundary, yielding a deterministic and controllable exploration procedure that both quantifies the extent of performance modes and identifies counterexamples without reliance on intermediate models.

\subsection{Contributions}

Building on our previous work \cite{VectorCostBimatrix}, we expand the functionality of the vector cost bimatrix game for increased generalizability. Evaluation of new cost structures to handle more complex planning scenarios relative to the scalarization approach is augmented with state of the art validation techniques and visualizations, such as SHAP feature importance and performance mode volume. 

Our major contributions in this paper include:
\begin{itemize}
    \item A systematic procedure with vector costs for robotic behavioral planning with arbitrary numbers of objectives.    \item XAI feature analysis for improved interpretation and analysis of agent decision making.
    \item SEMBAS-targeted sampling for extensive sensitivity analysis and performance evaluation. 
\end{itemize} 

This paper is organized as follows. We introduce the expanded vector cost bimatrix game in Sec.~\ref{sec:problem_formuation}. We describe the optimization algorithm used in the vector cost approach, details on cost design decisions, and the validation tools in Sec.~\ref{sec:methods}. Then, in Sec.~\ref{sec:results} we present the results of a comparative study on the performance of scalar and vector-based behavioral planning algorithms. 

\section{Problem Formulation}\label{sec:problem_formuation}


A vector cost bimatrix game has a two player structure. Both players seek to minimize the accrued costs as a result of their combined actions, though they do so independently and without direct cooperation. Each player has discrete actions that they choose out of actions sets $U_1:=\{1,\dots, n\}$ and $U_2 = \{1,\dots, m\}$ simultaneously. Their policies are denoted by $\gamma \in U_1$ for player 1 and $\sigma \in U_2$ for player 2. Each component cost matrix for the objective functions of the players is given by $C_i^h \in \mathbb{R}^{n \times m}, \forall h \in [1,\ldots,g]$. The vector of each player's cost functions is $C_i \in \mathbb{R}^{n \cdot g \times m}, \text{ for } i \in \{1,2\}$ and $g$ different objectives, and outcomes $J_i \in \mathbb{R}^{g \times 1}, \text{ for } i \in \{1,2\}$ . We organize these costs in order of importance, with the first objective captured by $C_i^1$ being the highest priority, and the last objective encompassed by $C_i^g$ being the lowest. The full structure of the game can be described as
\begin{equation*}\begin{aligned}
C_i=
\begin{bmatrix}
    C_i^1\\
    \vdots \\
    C_i^g, 
\end{bmatrix},
\quad J_i(\gamma,\sigma)=
\begin{bmatrix}
    C_i^1(\gamma,\sigma)\\
    \vdots \\
    C_i^g(\gamma,\sigma)
\end{bmatrix}\end{aligned}. \label{vector cost}\end{equation*}

The scalarization approach described in \cite{ScalarizingMultiObjective} can be taken as the creation of a weighted sum of objective functions $D_i \in \mathbb{R}^{n \times m}$. This is calculated from weight vector $\Theta_i \in \mathbb{R}^{1 \times g}$,
\begin{equation*}\begin{aligned}
\Theta_i=
\begin{bmatrix}
    \theta_i^1
    \ldots 
    \theta_i^g
\end{bmatrix}\end{aligned} \label{eq:weight vector}\end{equation*}
using the following summation:

\begin{equation}
    D_i(C_i,\Theta_i) = \sum_{h=1}^g \theta_i^h \cdot C_i^h.
\end{equation}

In order to scale to larger numbers of dimensions, we redefine the sets of Pareto optimal policies from \cite{VectorCostBimatrix} in a more compact form. These policies dominate all others with respect to at least one cost type and are no worse with respect to the other objectives. This and the following sets are calculated for a given fixed policy $\sigma$.
\begin{equation}\label{eq:triple pareto}
    \mathcal{P}_1(J_1(\gamma, \sigma)) = 
    \left\{
    \begin{aligned}
        \gamma^* \in U_1 \, : \, 
        J_1(\gamma^*,\sigma) \dot{\leq} J_1(\gamma,\sigma) \\
    \end{aligned}
    \right\}.
\end{equation}

Similarly, the set of policies that result in worst-case outcomes with respect to at least one objective is defined as:
\begin{equation}\label{eq:worst}
\mathcal{W}_1(J_1(\gamma, \sigma)) = \bigcup_{h=1}^g \argmax_{\gamma^w \in U_1}C_1^h(\gamma,\sigma)
\end{equation}

We use this to quantify robustness as a metric for evaluating the effectiveness of a given policy, which can be said to be the extent to which the policy $\gamma$ yields outcomes $J_1(\gamma,\sigma)$ that are not in the worst case set $\mathcal{W}_1$ regardless of input $\Theta_1$. 

While fixing player 2's policy as a security policy $\sigma^s$, we can describe a set of moderate Pareto optimal policies that are both optimal and avoid the worst case outcomes for all objectives. 
\begin{equation}\label{eq:moderate}
\mathcal{M}_1(J_1(\gamma, \sigma^s)) = 
\{ \gamma^m \in U_1 : \gamma \in \mathcal{P}_1\setminus \mathcal{W}_1\}.
\end{equation}

{\bf Problem statement:} For arbitrary input values $\Theta_1,\Theta_2$, we intend to design a composite cost matrix $\tilde{D}_1$ from $C_1$ in a manner that preserves a greater amount of structure from each objective function, a process we term the vector cost method. We do this by designing an adjustment matrix $E$ so that the resulting cost matrix $\tilde{D_1} = E+C_1^1$ produces a security policy $\tilde{\gamma}^s\in \mathcal{M}_1$.

Using the scalar cost bimatrix game as a baseline method, we would like to establish through this comparative study whether the vector cost method is more robust than its counterpart to suboptimal inputs of priority weights $\Theta_1$ for the ego agent. This may be in situations where the weights have not been tuned properly, environmental changes render the weights maladaptive, or the executive function of the vehicle is compromised by a disturbance. 

\section{Methods}\label{sec:methods}

\subsection{Cost Adjustment through Convex Optimization}

The design of the new cost matrices to force security policies that are both Pareto optimal and robust is guided by the concept of potential games. To find $E$ for the calculation of $\tilde{D_1}$, we produce a potential function $\phi \in \mathbb{R}^{n \times m}$ from the bimatrix pair $\tilde{D}_1, D_2$ that adheres to the exact potential game structure. The minimum entries of $\phi$ match with the Nash equilibria of the corresponding bimatrix game. When there is one unique global minimum of $\phi$, we would like the unique admissible Nash equilibrium to be equivalent to the pair of security policies $\{\tilde{\gamma}^s,\sigma^s\}$ for that same bimatrix game. The equilibrium solution to the game is the set of policies such that no player has incentive to deviate from their current strategy.
\begin{equation}
D_1(\gamma^*, \sigma^*) \leq 
D_1(\gamma,\sigma^*), \, 
D_2(\gamma^*, \sigma^*) \leq 
D_2(\gamma^*, \sigma),
\label{equilibrium} \end{equation}

We capture this in the following optimization. The input is the desired location of the minimum of $\phi$ given by entry $(r,c)$ and the cost matrices $C_1^1$ and $D_2$.  
\begin{equation}
\begin{aligned}
& \min_{E, \phi \,: \, \phi(r,c) = 0} && \|E\|_F^2\\[1ex]
& \text{subject to} && \forall \gamma, \bar{\gamma} \in U_1, \text{ and } \forall \sigma, \bar{\sigma} \in U_2, 
\\& && C_1^1(\gamma,\sigma)+E(\gamma,\sigma)-C_1^1(\widebar{\gamma},\sigma)-E(\widebar{\gamma},\sigma) =\\
& && \quad  \phi(\gamma,\sigma)-\phi(\widebar{\gamma},\sigma), \quad \\
& && D_2(\gamma,\sigma)-D_2(\gamma,\widebar{\sigma}) =\phi(\gamma,\sigma)-\phi(\gamma,\widebar{\sigma}), \\
& &&  \phi(\gamma,\sigma) > 0, \forall (\gamma, \sigma) \neq (r,c). 
\end{aligned}
\label{eq:error minimization}
\end{equation}
The first two constraints model the requirement that the matrix $\phi$ is a potential function for the bimatrix game $C_1^1+E, D_2$. The final constraint models the requirement that the global minimum of $\phi$ is exactly at the specified choice location $(r, c)$.

Algorithm~\ref{alg:policy select} computes the optimal error matrices for every choice of candidate global minima $(r,c)$ in $\phi$. Failure to find a suitable $E$ using any $(r,c)$ causes player 1 to default back to the scalarization method. One advantage of this algorithm is that the only required step to scale from two objectives to any number is to update the Pareto optimality calculation for each additional cost type. The adjustment algorithm will use the prime objective $C_1^1$ for the creation of the Nash equilibrium policy regardless of the number $g$ objectives.  

\begin{algorithm}
\caption{Policy Selection from Adjusted Costs}
\label{alg:policy select}
\begin{algorithmic}[1]
    
\State \textbf{Input:} Cost matrices $C_1, D_2$, Weights $\Theta_1,\Theta_2$ 
\State \textbf{Output:} Policies $\tilde{\gamma}^s, {\sigma}^s$ for players 1 and 2.
\State Compute security policy: ${\sigma}^s = \argmin \max D_2$
\State Initialize best error matrix: $E^* = \infty$
\For{$r$ in $\mathcal{M}_{1} (J_1(\gamma, \sigma^s))$}
    \State Find optimal $(E, \phi)$ via Equation~\eqref{eq:error minimization} using $(r, {\sigma}^s)$.
    \If{$E$ is finite \textbf{and} $\norm{E}^2_F < \norm{E^*}^2_F$}
        \State $E^* = E$
    \EndIf
\EndFor
\State Compute,\[ 
\tilde{\gamma}^s = \begin{cases} \arg\min \max ({C}_1^1 + E^*), &\text{ if $E^*$ is finite}, \\
\arg\min \max (D_1(C_1,\Theta_1)), &\text{ otherwise.}\end{cases}
\]
\end{algorithmic}
\end{algorithm}

\begin{theorem}[Output of Algorithm~\ref{alg:policy select}]\label{thm:1}
Given input matrices $C_1^1$ and $D_2$ and for any fixed set of weights $\Theta_1, \Theta_2$, the resulting policies $\{\tilde{\gamma}^s$, $\sigma^s\}$ form a pair of security policies for the bimatrix game $\tilde{D}_1, C_2$ and also a Nash equilibrium. 
\end{theorem}

\textbf{Proof.} This proof is similar to the case with two objectives, since we are still dealing with a single pair of cost matrices that form a potential function. We leverage the fact that all global minima of a bimatrix potential game correspond to Nash equilibria of the bimatrix game \cite{NoncooperativeGame}. The optimization in Equation~\eqref{eq:error minimization} finds $E$ that defines $\tilde{D}_1$ such that there is a unique global minimum. The column $c$ corresponding to the global minimum is selected to match the security policy $\sigma^s$ (line 6 of Algorithm~\ref{alg:policy select}). Finally, in line 9, we calculate $\tilde{\gamma^s}$ as the security policy using the best error matrix. Therefore, the Nash equilibrium policies are also equivalent to security policies $\{\tilde{\gamma^s},\sigma^s\}$.  \qed

\subsection{Conditions for Algorithm Convergence} 

Here we define the necessary topography of the cost functions in order for the optimization to be successful. We define pairwise column and row differences $\Delta^{c}(\cdot) \in \mathbb{R}^{t \times m}$, $\Delta^r(\cdot) \in \mathbb{R}^{n \times s}$, where $t:=({n \atop 2})$ and $s:=({m \atop 2})$, as matrices containing every combination of differences between elements in the argument matrices along each column and row, respectively.

The relationship between entries in each column $j$ and across each row $i$ of a matrix $X$ can be defined as
\begin{equation*}
\begin{aligned}
    \Delta_{j}^{c}(X_j)=[x_{ij}-x_{kj}\,\,  | \,\, \forall i, k \ \; 1 \leq i < k \leq n]\\
    \Delta_{i}^{r}(X_i)=[x_{ij}-x_{ik} | \forall j, k \; \; 1 \leq j < k \leq m].
\end{aligned}
\label{eq:difference}\end{equation*}
We can now describe the constraints of the exact potential as
\begin{equation*}
    \begin{bmatrix}
        \Delta^c(\phi) \\
        \Delta^r(\phi)'
    \end{bmatrix}
    =
    \begin{bmatrix}
        \Delta^c(\tilde{D}_1) \\
        \Delta^r(D_2)'
    \end{bmatrix}.
\label{pairwise potential}\end{equation*}

\begin{theorem}[Required Cost Conditions]\label{thm:2}
For a given value of the costs $D_2$ for player 2, Equation~\ref{eq:error minimization} produces finite $E$ such that bimatrix game $\tilde{D}_1, D_2$ produces potential $\phi$ with a global minimum at chosen position $(r,c)$ if fixed costs $D_2$ adhere to the following conditions. For entry $d_{rj} \in \Delta^r(D_2)$:
\begin{equation}\label{eq:cost conditions}
    \begin{aligned}
    & d_{rj}<0, \forall j < c, \quad d_{rj}>0, \forall j > c.
    \end{aligned}
\end{equation}
\end{theorem}

\textbf{Proof.} The potential game can be expressed only in terms of the fixed costs $D_2$ that are unaffected by the cost adjustment algorithm and determine the viability of certain selections of global minimum. The potential game constraints are now simply $\Delta^r(\phi)=\Delta^r(D_2)$, with $\Delta^c(\tilde{D}_1)$ as a free variable that can be adjusted to accommodate the pairwise row differences. When selecting a global minimum based on player 2's costs, the entries in the potential function must be of the form:
\begin{equation*}
    \phi = \begin{bmatrix}
        k_1-d_{11} & \ldots & k_1 & \ldots & k_1+d_{1g} \\
                \vdots &  &\vdots & &\vdots \\
        -d_{r1} & \ldots & 0 & \ldots & d_{rg} \\
        \vdots &  &\vdots & &\vdots \\
        k_n-d_{n1} & \ldots & k_n & \ldots & k_n+d_{ng}
    \end{bmatrix}.
\label{row differencce potential}\end{equation*}
Every entry is defined in terms of the pairwise difference of the row of the original costs and arbitrary offset constant $k_i>0$ for each $i$th row. This constant is selected so that $\phi(r,c)=0$ is the lowest value, which serves to enforce that minimum across all rows. However for row $r$, entries $d_{rj}$ must be already be in a range that results in positive values across every entry that is not the minimum. Therefore, in order to select global minimum $(r,c)$, the row pairwise differences of $D_2$ must adhere to the constraints described in Theorem~\ref{thm:2}.
\qed

\subsection{Cost Structure Design and Optimization}

We explore two different sets of cost functions and their effects on the performance of the scalar cost and vector cost approaches. We focus on a set of example cost functions here to illustrate the extension of the vector of costs to higher dimensions. The situation used in the game is a two player racing scenario on a circular road where both agents aim to get ahead of the other while staying on the track and avoiding collisions. Player 1 will start behind player 2, and we will designate it as the attacker. The attacker will be attempting to overtake player 2, the defender, in front of it. We choose to consider three different objectives for the remainder of the paper: cost structures related to the progress of the agent towards the goal, its adherence to staying within the bounds of the playing area, and the penalty for collision risk due to its proximity to the opponent.

\begin{figure}[h]
    \centering
    \includegraphics[width=1\linewidth]{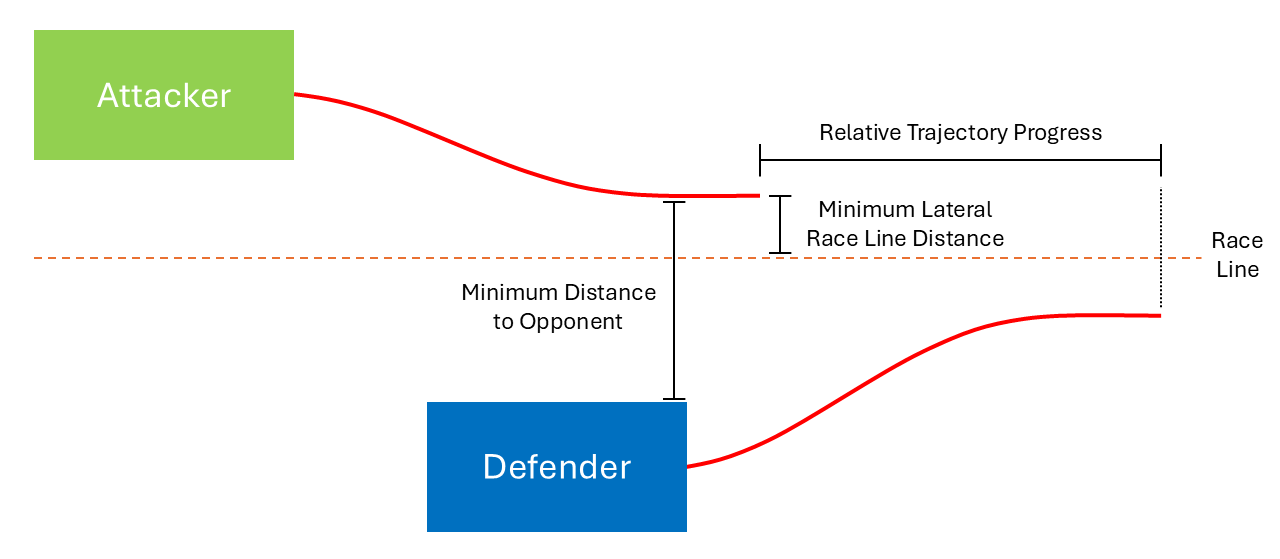}
    \caption{Distances used for cost design}
    \label{fig:trajectory_schematic}
\end{figure}

If we give each point in space a marginal cost when added to a trajectory, we can overlay the costs onto physical space and view their shape, much like control barrier functions. These point costs we can describe in each cost category as $c^1, c^2, c^3 \in \mathbb{R}$, the progress, boundary, and proximity costs of a single player. The physical basis of these quantities can be visualized in Figure \ref{fig:trajectory_schematic}.

A simple group of linear cost functions can be mapped onto the circular racing environment. The progress cost is calculated as a function of the relative angle between the two players around the circular track, $\alpha, \beta \in \{0,2\pi\}$ for players 1 and 2. The boundary cost is calculated as the Euclidean distance off the track, measured as the difference between the nearest boundary point $b \in \mathbb{R}^2$ and the point $p \in \mathbb{R}^2$. The proximity cost is calculated as the difference between the max value $q \in \mathbb{R}$ and the distance of the point to the opponent's position $ o\in \mathbb{R}^2$. This cost is only applied when the distance falls below the proximity threshold $\tau\in \mathbb{R}$.   

\begin{equation}\label{eq:first cost functions}
\begin{aligned}
    c^1 = \beta-\alpha, \quad c^2 = \|b-p\|, 
    \\ c^3 = ( q - \| o - p\| ) \cdot \mathbf{1}_{\| o - p\| < \tau}
\end{aligned}
\end{equation}

\begin{figure}
    \centering
    \includegraphics[width=\linewidth]{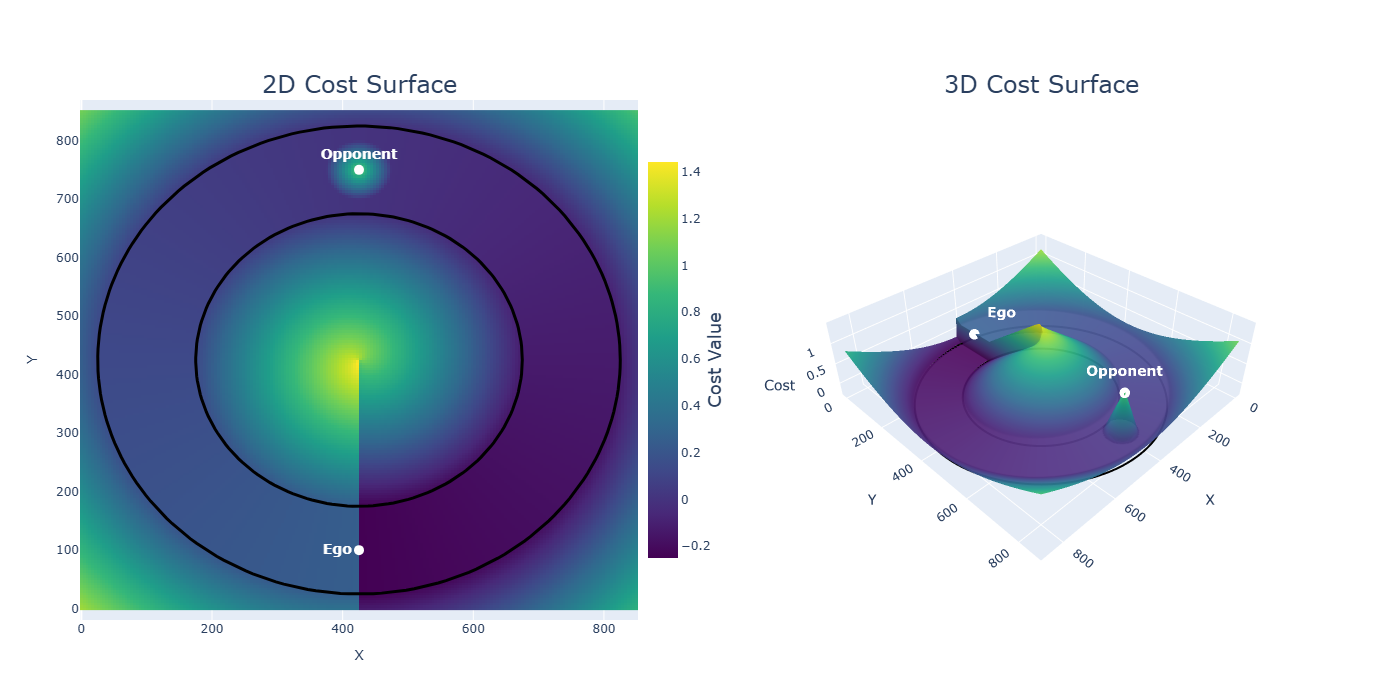}
    \caption{Linear cost structure}
    \label{fig:cost visual 1}
\end{figure}

In Figure \ref{fig:cost visual 1} we can see the costs projected onto the plane of the track itself. This is the linear sum of each cost type, scaled with $\Theta_i$ values that enable features from each category to be visible. The corkscrew pattern that slopes downward is the progress cost, which incentivizes motion in this direction. The boundary cost slopes upwards the farther the points get from the track, forming a valley to keep the players inside the track. Finally, a sharp peak can be seen around the opponent, forming a strong incentive to avoid collisions.

From the visual perspective, this combination of costs seems to encapsulate the goals of each objective. If we were to imagine the ego vehicle as a marble and dropped it into Figure \ref{fig:cost visual 1}, it would roll down the track in the proper direction, staying inside the boundaries and curving to avoid the opponent. 

A major weakness of these cost functions is also visible in the plot. Abrupt changes in the different regions where specific cost types become dominant pose a significant problem for the cost optimization algorithm. These linear functions tend to have derivatives that quickly flatten out, causing difficulties in convergence. This failure to converge often produces new cost matrices that satisfy the potential game constraints, but have not been developed enough to satisfy the requirements for matching security policies to the Nash Equilibrium. Specifically, the adjusted cost matrix $\tilde{D}_1$ will form potential matrix $\phi$ with opponent composite costs $D_2$ at the desired minimum position $(r,c)$, but policy $\tilde{\gamma}^s$ over $\tilde{D}_1$ will not match $r$ due to aberrant values in $\tilde{D}_1$. In practice, the algorithm filters out these undesirable solutions, though these conditions corrupt enough of the viable minima to cause reversion to the scalar method virtually every time. This was not the case for the application to two costs, where the optimization succeeded roughly half the time.

In order to produce conditions more favorable to convergence, we redesign the point costs as those in Equation \ref{eq:second cost functions}. We keep the progress costs identical since they produce gradual, smooth curves both in the base function and its derivatives. The bounds costs are now defined in terms of the distance of point $p$ from the closest centerline point $l \in \mathbb{R}^2$ and a slope tuning term $s_b \in \mathbb{R}$. The proximity costs are still defined by the distance to the opponent, now controlled by another slope term $s_c \in \mathbb{R}$.

\begin{equation}\label{eq:second cost functions}
\begin{aligned}
    c^1 = \beta-\alpha, \quad c^2 = 1-\exp \Big(\frac{-2}{s_b} \|l-p\|^2 \Big),
    \\ c^3 = \exp \Big(\frac{-2}{s_c}\| o - p\|^2 \Big)
\end{aligned}
\end{equation}

\begin{figure}
    \centering
    \includegraphics[width=\linewidth]{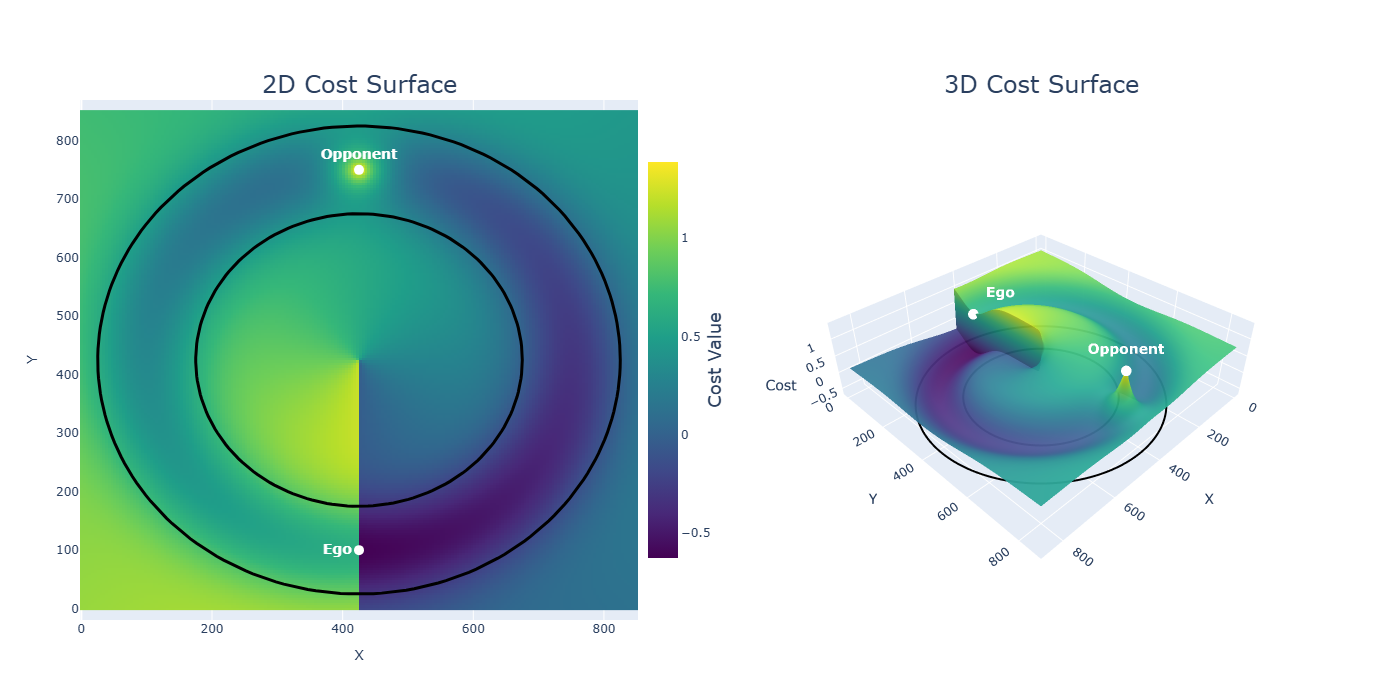}
    \caption{Radial basis cost structure}
    \label{fig:cost visual 2}
\end{figure}

The structure of these new cost functions closely resemble radial basis functions in the way they map the physical distances onto the cost structure. In Figure \ref{fig:cost visual 2} we see a surface that resembles Figure \ref{fig:cost visual 1}, but with important differences. The bounds and proximity cost regions transition to areas of higher influence gradually, contrasting to before when changes were more abrupt. The bounds cost in particular now extends onto the track itself, incentivizing the middle of the road as the primary race line. 

Another important property is that the bounds cost is nonzero for the region inside the track boundaries, which was not the case before. When a specific cost type has zero cost in a region, the trajectories plotted through that region are indistinguishable to each other by that cost type. Looking at the Pareto frontier yields a 2D plane in the other two dimensions when it should be a 3D surface. Since the Algorithm \ref{alg:policy select} searches for moderate Pareto optimal trajectories that are neither the minimum or maximum for any cost type, this automatically rules out all potential candidates.

While the new bounds cost avoids this, the proximity costs are still limited to the region around the opponent by design. The safety-critical decisions when the ego vehicle is interacting directly with the opponent can be made with the more robust but computationally intensive framework, while the majority of operations are handled by the simpler scalarization method. 

\begin{figure*}[!th]
    \centering
    \includegraphics[width=\linewidth]{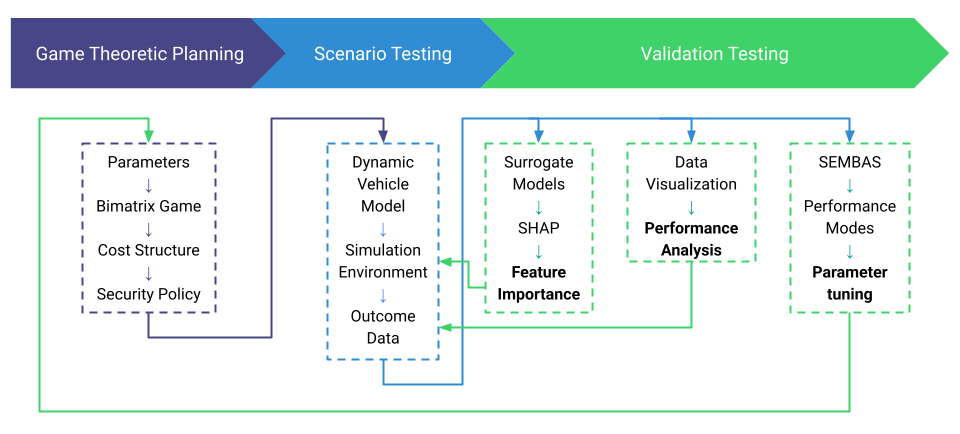}
    \caption{Simulation and validation pipeline}
    \label{fig:pipeline}
\end{figure*}

Note that for the two-cost case, the cost adjustment method operated only when safety costs were nonzero, namely when trajectories were projected to intersect with the track boundary or the path of the opponent. This made for more rounds with viable candidate minima, while for the three-cost case there were fewer turns with successful optimizations but more candidates to choose from in the set $\mathcal{M}_1$. This provides an intuition as to how the vector cost approach scales to higher dimensions. Flat or sparse regions within the hypervolume of the cost space will not yield favorable conditions for the algorithm, while critical areas with meaningful topography will provide ample opportunities for mechanism design.

\subsection{Validation Tools and Algorithm Improvement Pipeline}

In addition to traditional quantitative analysis techniques, we make use of several types of state-of-the-art validation techniques. 
%
This includes the tools broadly included under the umbrella of XAI. After inputting a large amount of tabular data from simulation into the pipeline, we can generate surrogate models that approximate the input/output relationship of the data. These models can give rapid output of additional data without having to repeat tests with the simulation in the loop. This is important for the process of Shapley Additive Explanations (SHAP), an approach that determines the impact of a given feature on another target value \cite{AUnified}. The performance of the models is evaluated for many different combinations of feature inputs, then the degree to which a feature's absence affects the prediction accuracy is used to determine how important it is to the overall output. Using these rankings we can better interpret the choices of the agents and the impact of the simulation parameters on their actions.

Finding useful test scenarios and analyzing their results can be challenging in continuous, higher dimensional parameter spaces. With many systems have trivial extremes, such as impossibly difficult driving conditions, it is important to focus testing on the boundary of performance modes. State-space Exploration of Multidimensional Boundaries using Adherence Strategies (SEMBAS) \cite{SEMBAS} provides tools for exactly this purpose. SEMBAS is a library that includes algorithms for efficiently exploring parameter spaces, identifying performance boundaries, and exploring those boundaries. The output includes a geometric representation of the boundary, detailing where the boundary is located and which direction it faces. Consequently, this boundary can be used to measure its internal volume, acting as an estimate for how likely a given performance mode is to occur. These boundary cases can be manually analyzed to verify that the system is performing as expected.

The relationship between these tools is described in Figure~\ref{fig:pipeline}. The planning algorithms are implemented in the bimatrix game style with the scalar and vector cost structures. The setup of our simulations is captured in the scenario testing framework where we map the decision making onto the dynamics of the vehicles, place them in the race track environment, and gather data from their interactions. The validation testing makes use of this data in a number of ways.

Surrogate models are trained from the data and used to quantify what physical or game theoretic elements have the greatest impact on decision making. The tools built into the XAI pipeline allow for easy representation of this data to analyze performance and identify missing scenarios for testing coverage. SEMBAS's targeted sampling of the hyperparameter space also allows for the identification of defects, the determination of valid parameters that are effective and safe, and comparison of algorithms by measuring the size of the success parameter envelope.  

\section{Results}\label{sec:results}

In this section we evaluate the performance of the scalar and vector cost methods using data obtained from a custom simulation. We use a traditional grid search to explore the cost weight parameter space and apply SHAP value feature analysis to evaluate the effect of state variables and cost structures on decision-making performance. As a counterpoint, we use SEMBAS for targeted sampling of the same parameter range to find performance mode boundaries and quantify robustness.

\subsection{Tabular data and Feature Analysis}

We constructed our racing simulator using the kinematic bicycle model to approximate vehicle movements \cite{PotentialGameBased}. Here the dimensions of the simulated cars are 4 meters in length by 2 meters in width. The race course has an outer radius of 40 meters and an inner radius of 25 meters for comparison. The two agents are classified as defender and attacker. The defender always spawns in front of the attacker on the circular race track, and the attacker must pick actions such that it is able to overtake the defender. To facilitate this, the attacker is given a 50\% greater maximum speed. Both agents have the same action space, 9 different choices of static trajectories representing different combinations of acceleration and steering angle. The players pick their trajectories on each turn by playing the vector cost bimatrix game repeatedly over 30 rounds, long enough for the attacker to pass the defender should it make the proper set of choices. We use the cost structure described in Equation~\eqref{eq:second cost functions} to encode the three major objectives.

\begin{figure}[t]
    \centering
    \subfloat[Close Tail]{%
        \includegraphics[width=0.45\linewidth]{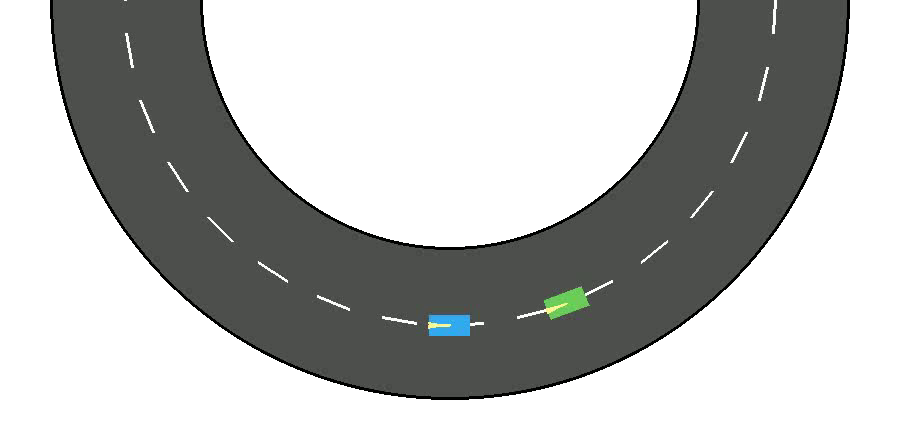}}
    \hfill
    \subfloat[Far Tail]{%
        \includegraphics[width=0.45\linewidth]{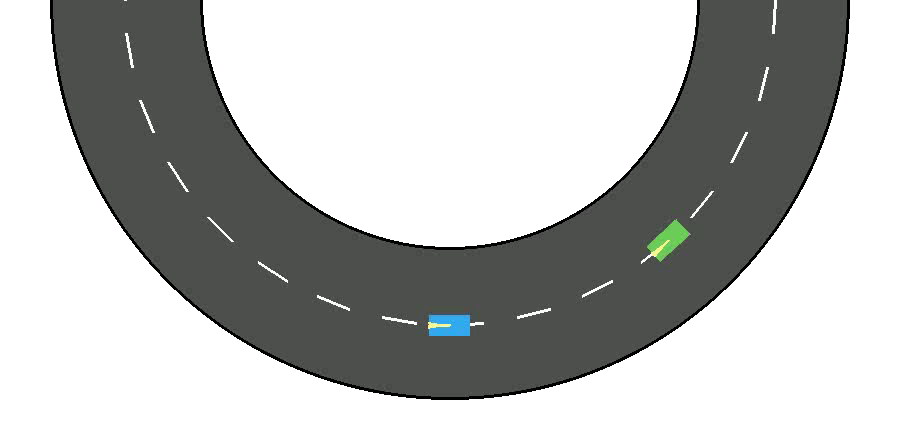}}

    \vskip\baselineskip

    \subfloat[Inside Edge]{%
        \includegraphics[width=0.45\linewidth]{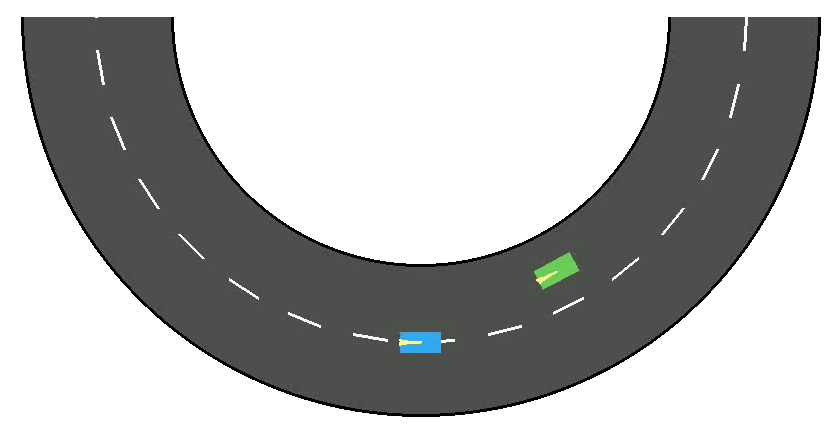}}
    \hfill
    \subfloat[Outside Edge]{%
        \includegraphics[width=0.45\linewidth]{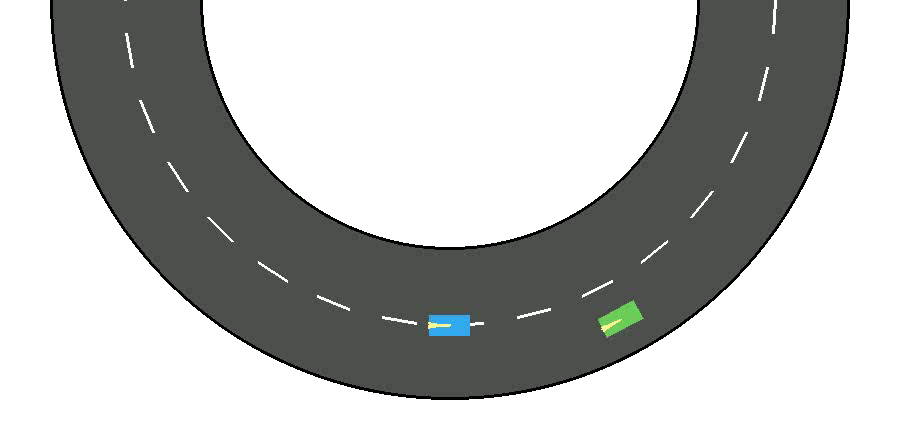}}

    \caption{Spawning scenarios for grid search testing}
    \label{fig:spawnpoints}
\end{figure}

To analyze the effectiveness of the scalar and vector cost structures, we ran a total of 500 races for each method. Each was tested on four different initial configurations that changed the starting position of the attacker behind the defender, varying both the distance between the players along the centerline of the track and the lateral position of the attacker relative to the centerline. See Figure~\ref{fig:spawnpoints} for visuals of these scenarios.

In addition, for each race we selected a different combination of $\Theta_1$ values for the attacker in the range $\theta^1_1 \in [0.001,1]$ for the progress cost weight and $\theta^2_1, \theta_2^3 \in [0,1]$ for the bounds and proximity cost weights. This was to evaluate how the weight configuration would affect the agent's ability to balance all three objectives at once. Note that we did not include the case of $\theta^1_1=0$ because this would eliminate the attacker's motivation to move at all. Five different intervals of these ranges were used for each weight parameter, for a total of 125 different weight combinations across 4 initial position scenarios. The configuration of the defender was held constant at $\Theta_2=[1,1,1]$, an equal consideration of all objectives.  

Table \ref{tab:race_stats} shows the raw statistics across all the races in the grid search. In general, the vector cost method outperformed the scalar cost method in passing effectiveness and adherence to the track boundaries. Notably, neither method engaged in maneuvers that caused collisions in any of the tested scenarios or weight configurations. While this bodes well for their collective safety record, this makes it difficult to compare the agents' relative effectiveness in this manner. To address this, we add the minimum distance measurement as a proxy for collisions. The vector cost method observes a greater average minimum distance, meaning that the agent was able to give the opponent a wider birth while executing maneuvers successfully. In addition, we tracked the states and cost matrix entries of both vehicles for each round of the race, totaling roughly 15,000 features.

\begin{table}[t]
    \caption{Attacker Racing Performance}
    \centering
    \begin{tabular}{lr@{}lr@{}l}
    \toprule
    \textbf{Algorithm} & \multicolumn{2}{c}{\textbf{Scalar}} & \multicolumn{2}{c}{\textbf{Vector}}\\\midrule
    Passes                   & 318 &.& 448 &.  \\
    Out of Bounds            & 120 &.& 50 &.  \\
    Collisions               & 0 &.& 0 &. \\
    Average Minimum Distance (m) & 2&.6 & 3&.9 \\
    Average Progress Cost    & 7&.09 & 2&.30 \\
    Average Bounds Cost      & 4&.35 & 3&.91 \\
    Average Proximity Cost   & 0&.76 & 0&.86 \\
    Proportion in Lead (\%)  & 30&.42 & 42&.28  \\
    Time Complexity          & $O(n$&$mg)$ & $O(n^2m$&$+nmg)$ \\
    \bottomrule
    \end{tabular}\label{tab:race_stats}
    \par\vspace{2mm}
    \parbox{0.01\linewidth}{\footnotesize}  Note: Complexity with action space sizes $n,m$ and $g$ objectives.
\end{table}

\begin{figure}[t]
    \centering
    \includegraphics[width=1\linewidth]{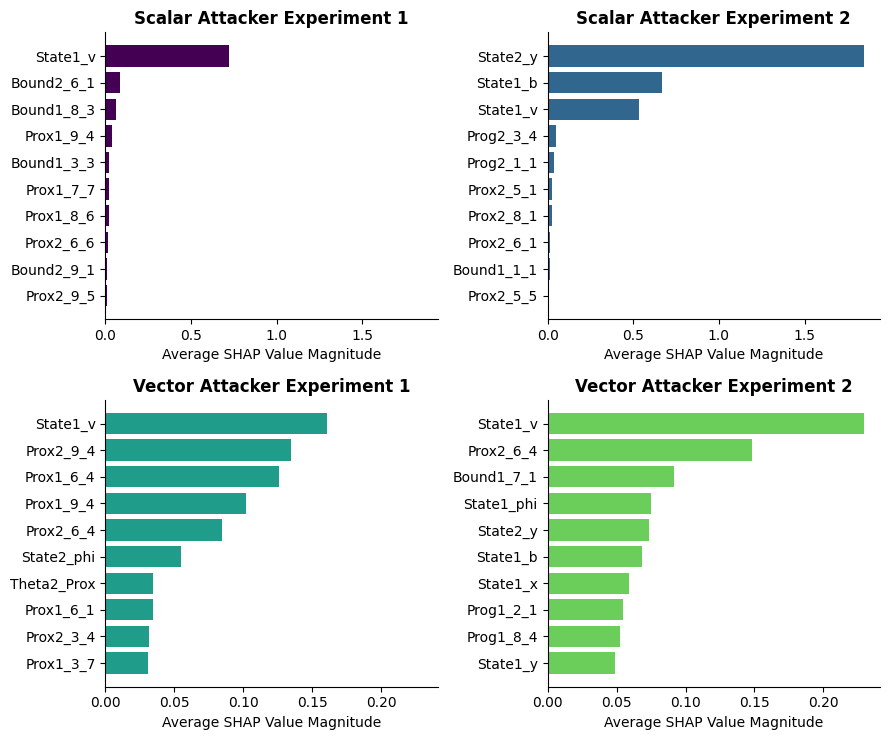}
    \caption{Feature importance to attacker actions in varying spawn conditions}
    \label{fig:action_shap}
\end{figure}

While these high level performance metrics are useful for comparison, diagnosing why a particular method performed better in a specific scenario requires a different set of tools. Evaluating the generalizability of the chosen scenarios and their effectiveness at gathering information from the systems under test is more subtle. In Figure \ref{fig:action_shap} we show the results of SHAP feature importance analysis on two different experiments. Experiment 1 was conducted identically to what was described above, except instead of using the four spawning scenarios, it included only cases with the two vehicles spawning along the centerline of the race track. Experiment 2 used the four spawning scenarios. The average magnitude of SHAP feature importance was measured with the target variable of the chosen actions of the attacker at each round. These importance values indicate a correlation between these particular features and the specific trajectories selected at any given time.

We measured the distribution of feature importance using Shannon entropy in order to quantify the degree to which the impact was spread to many factors or concentrated among some specific variables. A low entropy score indicates a more concentrated distribution, while high entropy is measured for data with large distributions \cite{ShapGuided}. The Shannon entropy was measured as 1.233 nats for the scalar attacker in Experiment 1 and 1.409 nats for the same in Experiment 2. The vector cost attacker scored 4.516 nats for its own actions in Experiment 1 and 3.924 nats in Experiment 2.

One interpretation of the entropy measurements is that a higher nat value indicates a higher robustness since the output of the agent depends less on any particular feature. This would suggest that an outsized disturbance to any particular feature would have less of an effect should the feature importance be shared more evenly. From Table ~\ref{tab:race_stats} we see that the vector cost method outperformed the scalar metric in all three major categories while varying the spawn point and the weight parameters, meaning that it is more robust with respect to these inputs in the most common scenarios. Looking at the feature importance distributions, we see that the vector cost method has roughly 3-4 times the Shannon entropy compared to the scalar cost method, which further strengthens this claim.

If we compare the entropy values across the experiments, the results are less straight-forward. Experiment 2 introduces variance in the attacker spawn point along the lateral and longitudinal axes, while Experiment 1 only changes the longitudinal distance relative to the defender. We would expect there to be an increase in the entropy between these two experimental setups. The scalar cost method follows this pattern with a higher entropy for Experiment 2 compared to Experiment 1, but the variance in the distribution of the vector cost feature importance actually decreases when adding the lateral movement. This suggests that the relationship between robustness and the entropy of the feature importance distributions is more complex than our initial hypothesis.

 The notation associated with each feature is structured as Name\_i\_n\_m\_r, where i refers to player 1, the attacker or player 2, the defender. The subscripts n, m, only pertain to cost matrix entries and are the row and column coordinates of that entry. The last number r indicates which round or decision epoch the reading was recorded and can be taken as a time-based parameter, where the duration of a single race is 30 rounds. For the cost types, we chose to abbreviate the progress cost as Prog, the bounds cost as Bound, and the proximity cost as Prox. The state variables included in these features include $x,y, v, \phi, \beta$, the first three being the positions and speed of the vehicle and the later two being the heading and lateral wheel slip.
 
 Looking at the individual feature importances, we can see that the velocity of the attacker when making a decision, State1\_v, has a high value in every case. This is especially true for vector attacker Experiment 2, where this particular feature is roughly 25\% greater than the next biggest feature Prox2\_6\_4, which is the cost value of the defender's proximity cost matrix at entry $(6,4)$. It follows that the velocity of the attacker would have a disproportionate impact on actions chosen since extreme values of speed, either very slow or very fast at the edges of the permitted range, are likely to cause the agent to choose to speed up or slow down in order to moderate being too far behind the defender or the execution of riskier high-speed maneuvers. 


\begin{remark}
    One major use of SHAP value analysis for experimental design might include in the evaluation of the chosen test scenarios themselves. If they effectively stress test the methods in question and contain enough variance in the independent variables, the features related to the chosen inputs will exhibit a higher feature importance. This is one concrete way to measure the quality of a scenario set and compare it directly to another while making test design decisions.
\end{remark}

\begin{figure}[t]
    \centering
    \includegraphics[width=\linewidth]{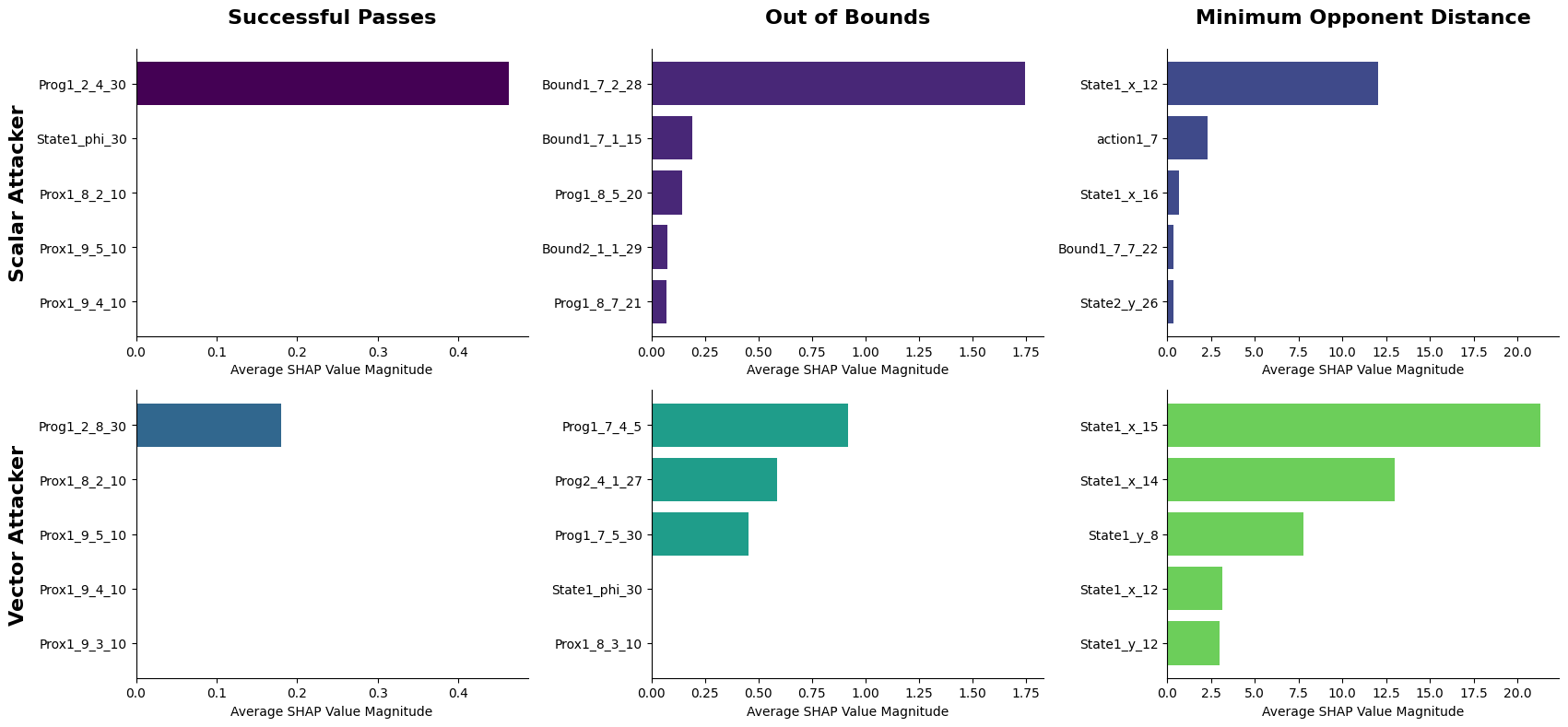}
    \caption{Feature importance with respect to race objectives}
    \label{fig:metric_shap}
\end{figure}

Figure \ref{fig:metric_shap} shows the feature importance of an expanded set of variables on the three key metrics from the racing experiments, successful passes, going out of bounds, and proximity to the opponent. The SHAP importance value can be biased at times towards specific features. For instance, in successful passes, both cost structures have the feature importance dominated by the feature Prog1\_2\_4\_30 and Prog1\_2\_8\_30, particular entries of the attacker's progress cost matrix at round 30 of the race. It is not reasonable to conclude that this is the only factor that influences whether or not the attacker conducts a successful pass. It is likely that this feature is especially large when the attacker has failed to pass, and small when the pass was initiated. The correlation was strong enough that it completely eclipsed all the other potential importance measurements.

The out of bounds column displays high importance in the area of mid to late stage bounds and progress costs for the scalar attacker, indicating that a major component of off the track incidents was the conflict between the passing and bounds goals in the later half of the race during rounds 15-30. In general the vector cost attacker had less of these conflicts, and only progress cost features showed up in the SHAP analysis. The fact that bounds costs did not appear as important variables suggests that the agent likely made choices that led to off the track incidents from positions close to the centerline, leaving the track quickly enough that it did not have time to consider the bounds costs at all. 

The final metric uses minimum distance to the opponent to measure risk of collision. No actual collisions were observed in the grid search data, so this intermediate metric was crafted as a means to determine how well the vehicles managed their proximity, with a lower minimum distance without collision indicating both higher motion planning precision and higher tolerance for collision risk. This metric is interesting because it sports the highest number of state variables, as opposed to cost variables that are effectively designed features that are a function of those states. The time aspect of these features has information about the portions of the race when the vehicles were in the closest proximity, with the scalar cost attacker drawing close to the opponent in rounds 12-22, slightly later than the vector cost attacker in rounds 8-15. 

\begin{remark}
    Feature importance can be used for temporal analysis as well. Identifying not just what scenarios and metrics are critical for evaluating a system but \emph{when} systems tend to succeed or fail can be instructive. This could be extended towards informing changes in design to improve performance at important moments, or conversely, towards deploying tailored disturbances to hamper an agent when they are most vulnerable.
\end{remark}

\begin{figure}
    \centering
    \includegraphics[width=1\linewidth]{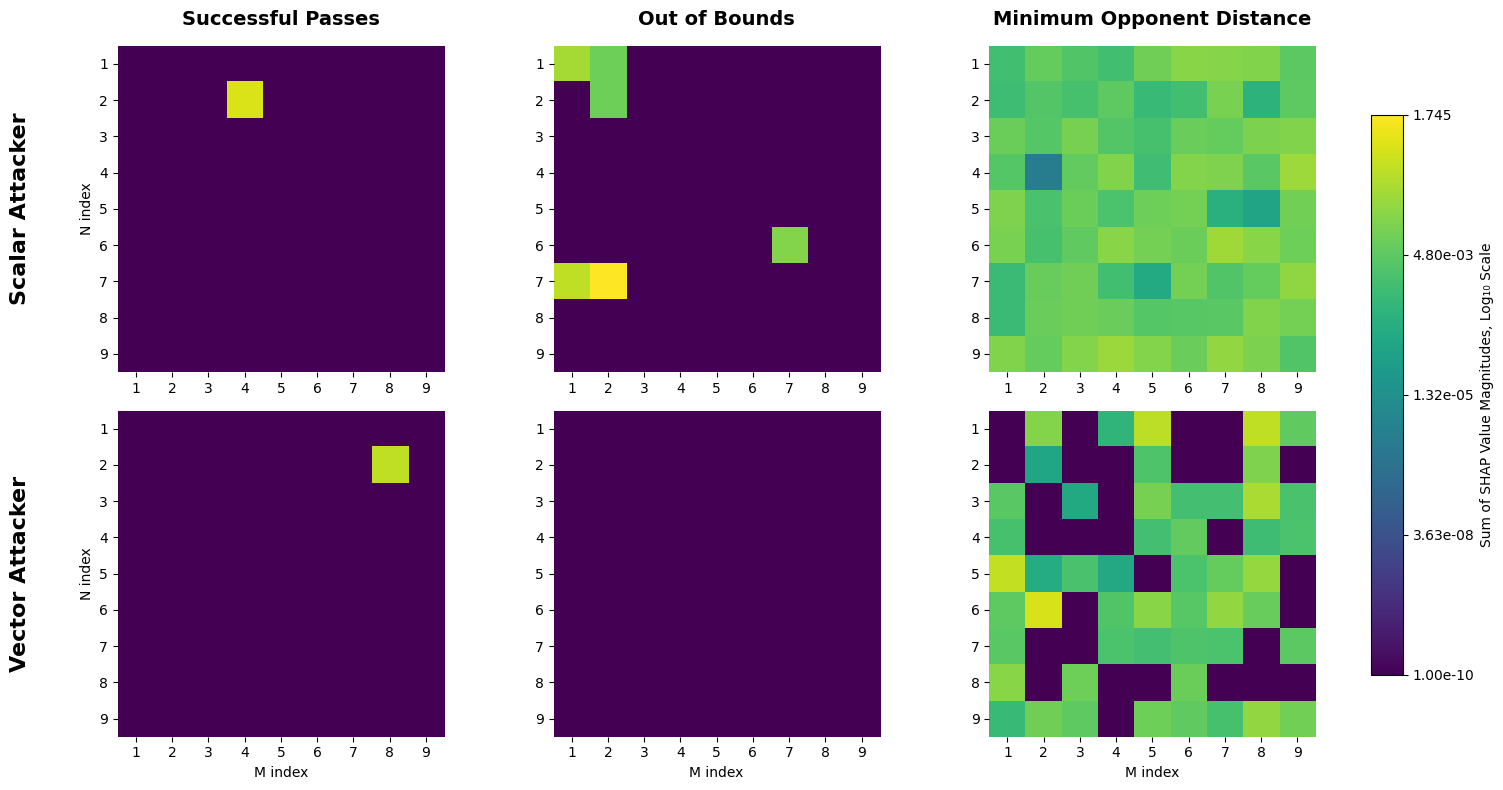}
    \caption{Feature Importance Heatmap of Cost Matrix Entries}
    \label{fig:cost_heatmap}
\end{figure}

A more detailed look at the importance of specific entries of the cost matrices can yield insights into both the cost structure and the action spaces of both agents. Figure \ref{fig:cost_heatmap} visualizes how important the outcome value at a particular intersection of actions was to determining the agent's success at each racing goal. Note for this visualization that the attacker is the row player who chooses index $n$ and the defender is the column player choosing index $m$. The importance is represented as the log SHAP value in this case since a linear scale would allow for only the entries with the most concentrated importance values to be visible.

Additional intuition is required for interpreting the indices of the cost matrices. The chosen action indexes the action space of trajectories made by nine different combinations of acceleration and steering angle, which can be divided into groups of three. Indices 1-3 are the trajectories that "slow down," each having a negative acceleration and steering angles for turning left, going straight, and turning right, respectively. Indices 4-6 have zero acceleration and 7-9 have positive acceleration, each with the same order of steering angles. 

With this in mind, we can interpret the importance in individual entries. For the passing metric, the most important indices were $(2,4)$ and $(2,8)$ for the scalar and vector methods. These correspond to the attacker choosing to slow down and go straight, while the defender maintains speed and turns left or speeds up and goes straight. We saw in Figure \ref{fig:metric_shap} that these importance values are for the cost matrices generated in the final round of the racing game. This could be a scenario where the defender is performing a maneuver in front of the attacker, who must slow down to avoid collision, but looking at the larger distribution of importance in this case still suggests an outlier here. 

The out of bounds metric has several entries that stand out in the scalar attacker's bounds cost matrix, while the cost values in the vector cost bounds matrix did not have a direct measurable effect. This is not to say that the vector cost method did not use the bounds cost for its decision making, more that a change to these values would not effect the outcome in a significant way compared to other factors. For the scalar attacker, the most important intersection of actions was $(7,2)$, when the attacker chooses to speed up and turn left while the defender slows down and goes straight. This corresponds to the scenario when the attacker is initiating the pass and moves to the outside of the track while the defender chooses to slow down to let it pass (or not). 
The aggressiveness of the pass is often moderated by the bounds constraint, which causes the attacker to moderate its speed in subsequent rounds, corresponding to the cluster of reduced acceleration entries $(1,1)$, $(1,2)$, and $(2,2)$.

The minimum opponent distance heatmap has the most interesting topography by far, likely because this is a continuous distance value while the other two are binary flags. The importance values are much more distributed for the scalar attacker, with most combinations of actions holding some significance, while the vector attacker observes more concentrated regions of importance. Here we can also gain information about what combinations of actions can be discounted for their effect in how close the agents become. For instance, there are the most low importance entries in rows 2 and 8, when the attacker is changing speed while not turning. This suggests that the opponents are closest together when this is not the case, when attacker is maintaining its maximum speed during a maneuver, or slowing down and turning to avoid a collision. 

\begin{remark}
This detailed information about the importance of different combinations of actions could be used for redesigning the action space of an agent to enable new behavior. For instance, if a specific index is more important in the cost matrix, more trajectories with slightly varied accelerations and steering angles that increase the granularity of choices in that area could be added. This could boost the amount of control the agent has over that outcome by giving it a better set of actions or tools to accomplish the goal.
\end{remark}

\subsection{Weight Sensitivity Analysis}

Grid search data gives an effective picture of the agent's performance over the entire parameter space, but with a limited resolution that is likely to miss many scenarios. SEMBAS uses boundary exploration to extract configurations that are exclusively edge cases, or scenarios that lie on the threshold between success and failure. Here we expand on our earlier analysis while focusing on the instances when the behavioral planning algorithms were pushed to the outer limits of their effectiveness.  

Instead of using weight values selected from a set of predetermined combinations, we connected the SEMBAS tool to the racing simulator using its websocket interface and had it select the three weight parameters for us. The SEMBAS algorithm then sampled race scenarios for each of the major spawning configurations and the two attacker cost structures, marking each race as a success or failure depending on whether the attacker was able to pass, stay on the track, or avoid collisions. After a brief global search period, the search algorithm was able to isolate a boundary surface that defined regions of the parameter space where the agent failed. The algorithm continued to sample along the boundary until 250 points were discovered or until the entire boundary was saturated to a resolution of 0.07.

Table~\ref{tab:sembas} shows the results of the SEMBAS search for all scenarios and objectives. In general, we see that the vector cost attacker had a higher estimated performance mode volume and a greater proportion of the parameter space for which the agent was successful. This region is normalized to a volume of 1. We observed that there was consistently one region of the weight value parameter space where failures occurred, showing a single cluster of unfavorable scenarios across all tests. Note that the entries that indicate a performance mode volume of 1.0 were scenarios in which SEMBAS was not able to find the boundary of a failure region within the maximum number of 500 samples. In comparing the robustness of these cost structures to maladaptive weight tuning, we can take these measurements as proof of the vector cost method's greater ability to avoid worst-case outcomes in the face of aberrant weight configurations.  

\begin{table}[h]
    \caption{Estimated Successful Performance Mode Volume}
    \centering
    \resizebox{0.47\textwidth}{!}{%
    \begin{tabular}{lr@{.}lr@{.}lr@{.}lr@{.}lr@{.}lr@{.}l}
    \toprule
    & \multicolumn{4}{c}{\textbf{Passes}} & \multicolumn{4}{c}{\textbf{Bounds}} & \multicolumn{4}{c}{\textbf{Collisions}} \\
    \textbf{Scenario} & \multicolumn{2}{c}{\textbf{Scalar}} & \multicolumn{2}{c}{\textbf{Vector}} & \multicolumn{2}{c}{\textbf{Scalar}} & \multicolumn{2}{c}{\textbf{Vector}} & \multicolumn{2}{c}{\textbf{Scalar}} & \multicolumn{2}{c}{\textbf{Vector}} \\\midrule
    Close Tail   & 0&671 & 0&866 & 0&985 & 1&0    & 0&942 & 1&0  \\
    Far Tail     & 0&510 & 0&963 & 0&935 & 0&985 & 0&934 & 1&0   \\
    Inside Edge  & 0&472 & 0&920 & 0&961 & 0&978 & 0&923 & 0&974 \\
    Outside Edge & 0&682 & 0&995 & 1&0    & 1&0    & 0&909 & 1&0   \\
    \bottomrule
    \end{tabular}\label{tab:sembas}
    }
\end{table}

\begin{figure}[h]
    \centering
    \includegraphics[width=1\linewidth]{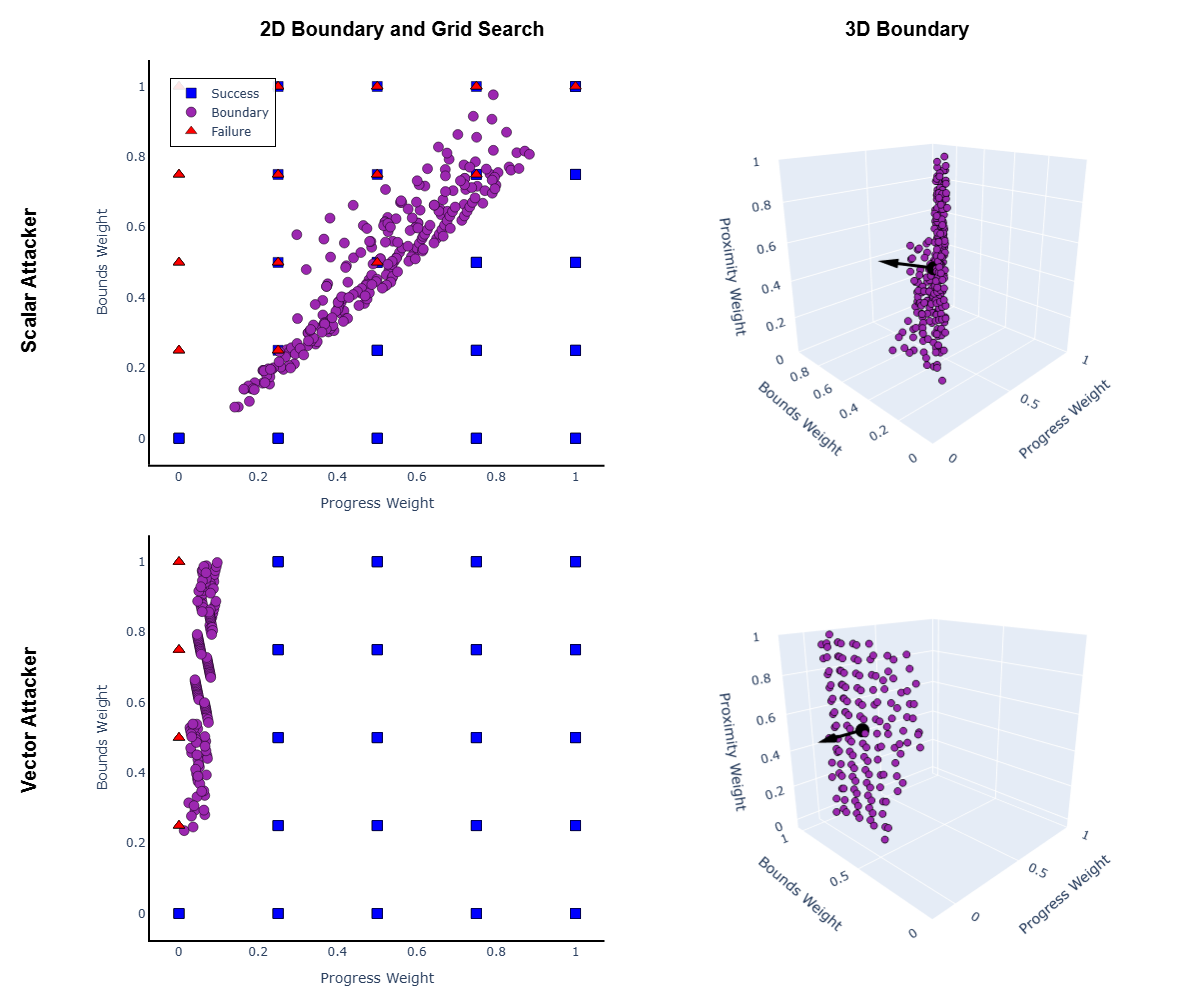}
    \caption{Grid search and performance mode boundaries in the space of cost weight parameters}
    \label{fig:grid_boundary}
\end{figure}

Figure~\ref{fig:grid_boundary} shows an example boundary in 2D and 3D views. This scenario is the inside edge spawn point, with successful passing as the target metric. The scatter plots in the left column include outcomes from both the grid search experiments and sampling on the boundary. The proximity weight dimension is omitted for the 2D view since it had a limited effect on the race outcomes for this scenario. For simplicity, only the boundary itself is shown in the 3D plots, with a red normal vector indicating the side corresponding to the failure performance mode. As expected, the boundary marks the transition between weight configurations that enable a successful pass and those that cause the agent to fail. Notice that the scalar cost attacker has a boundary that roughly bisects the parameter space, while the vector cost attacker must have its progress weight reduced to almost zero before it begins to fail.

\begin{figure}
    \centering
    \includegraphics[width=\linewidth]{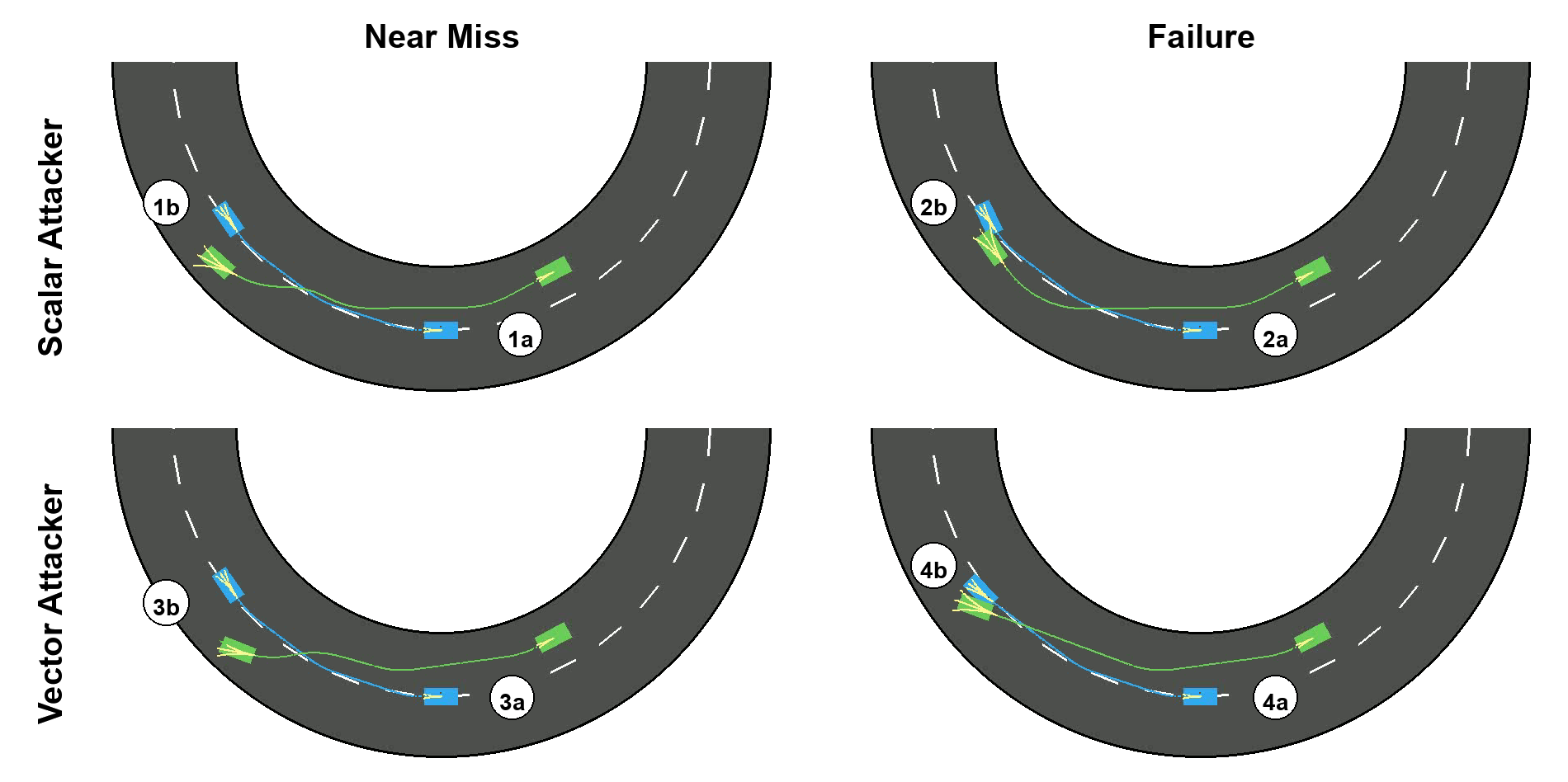}
    \caption{Key moments from collision avoidance edge cases}
    \label{fig:edge_case_comparison}
\end{figure}

Another benefit of the boundary adherence strategy is that the examination of the boundary configurations themselves can yield a rich set of examples for analysis and even for model training purposes. In Figure~\ref{fig:edge_case_comparison}, we show an overlay of key moments from a chosen set of boundary scenarios, two successful near misses, and two complete failures. Here we look at an inside edge spawn with collision avoidance as the metric. During the failures in scenes (2b) and (4b), the selected weight parameters cause the attacker to collide with the defender, roughly at the same time in the race for both methods. In scenes (1b) and (3b), slightly different parameter sets prompted the attacker to move farther to the outside before initiating the pass and completing the maneuver safely. By comparing these edge cases before and after crossing the boundary, we can also see that the actions of the attacker that led to success or failure for both methods are quite similar. The major difference in performance lies in the regions of the parameter space where these behaviors occur.

\begin{remark}
    Grid search and boundary exploration techniques are most useful when applied in tandem. The first of these measures the performance of the agent in the average case across many regions of the parameter space. However, this is insufficient for making generalizations about whether it is robust since we need to evaluate the system when subjected to extreme scenarios. Comparing the relative sizes of agent performance modes using a tool like SEMBAS is a more effective way to evaluate robustness.   
\end{remark}

\section{Conclusion}\label{sec:conclusion}

In this work, we expanded the vector cost bimatrix game for robotic behavioral planning applications with any number of objectives. We explored methods for mechanism design favorable to the cost optimization algorithm that is the core of the approach with an eye towards high dimensional performance. Simulations involving two agents in an overtaking race scenario were used to validate the effectiveness of both the scalar and vector cost approaches using several state of the art tools and metrics. SHAP feature analysis gave a comprehensive breakdown of the factors that affected the performance of the attacking agent, while SEMBAS provided a detailed view of the performance modes within the cost weight parameter space. Using these methods, we thoroughly validated our novel vector cost technique and showed that it outperformed baseline scalar cost methods.

Opportunities for future work are considerable. The vector cost bimatrix game has yet to be scaled to more than two players, and it could easily be adapted to multi-agent coordination scenarios where objectives are less competitive. The uses for XAI, boundary adherence algorithms, and other validation tools in the context of game theoretic agents are numerous and represent promising grounds for further study. 


\bibliographystyle{IEEEtran}
\bibliography{main.bib}

\end{document}